\documentclass[letterpaper, 10 pt, conference]{ieeeconf}  

\IEEEoverridecommandlockouts                              

\usepackage{graphics} 
\usepackage{epsfig} 
\usepackage{mathptmx} 
\usepackage{times} 
\usepackage{amsmath} 
\usepackage{amssymb}  

\usepackage{booktabs}
\usepackage{hhline}
\usepackage{color}
\usepackage{mathrsfs}
\usepackage{algpseudocode}
\usepackage{multirow}
\usepackage{caption}
\usepackage{subcaption}
\usepackage{diagbox}
\usepackage{bm}

\usepackage[pagebackref=true,breaklinks=true,letterpaper=true,colorlinks,bookmarks=false]{hyperref}

\title{\LARGE \bf
Realistic Rainy Weather Simulation for LiDARs in CARLA Simulator
}

\author{Donglin Yang, Zhenfeng Liu, Wentao Jiang, Guohang Yan, Xing Gao, Botian Shi, Si Liu, Xinyu Cai$^{*}$
\thanks{$^{*}$ Corresponding author.}
\thanks{Donglin Yang, Guohang Yan, Xing Gao, Botian Shi and Xinyu Cai are with Autonomous Driving Group, Shanghai AI Laboratory, China. {\tt\small \{yangdonglin, yanguohang, gaoxing, shibotian, caixinyu\}@pjlab.org.cn}
}
\thanks{Zhenfeng Liu is with Nankai University. {\tt\small 2120220408@mail.nankai.edu.cn}
}
\thanks{Wentao Jiang, and Si Liu are with Institute of Artificial Intelligence, Beihang University. {\tt\small \{jiangwentao, liusi\}@buaa.edu.cn} }
}

\begin{document}

\maketitle
\thispagestyle{empty}
\pagestyle{empty}

\begin{abstract}
Employing data augmentation methods to enhance perception performance in adverse weather has attracted considerable attention recently. Most of the LiDAR augmentation methods post-process the existing dataset by physics-based models or machine-learning methods. 
However, due to the limited environmental annotations and the fixed vehicle trajectories in the existing dataset, it is challenging to edit the scene and expand the diversity of traffic flow and scenario.

To this end, we propose a simulator-based physical modeling approach to augment LiDAR data in rainy weather in order to improve the perception performance of LiDAR in this scenario. We complete the modeling task of the rainy weather in the CARLA simulator and establish a pipeline for LiDAR data collection. In particular, we pay special attention to the spray and splash rolled up by the wheels of surrounding vehicles in rain and complete the simulation of this special scenario through the Spray Emitter method we developed. In addition, we examine the influence of different weather conditions on the intensity of the LiDAR echo, develop a prediction network for the intensity of the LiDAR echo, and complete the simulation of 4-feat LiDAR point cloud data. In the experiment, we observe that the model augmented by the synthetic data improves the object detection task's performance in the rainy sequence of the Waymo Open Dataset. Both the code and the dataset will be made publicly available at \url{https://github.com/PJLab-ADG/PCSim#rainypcsim}.

\end{abstract}

\section{Introduction}
In adverse weather, the perception performance of various autonomous driving sensors is negatively affected more or less. Limited by LiDAR's working principle, the effective detection distance of LiDAR decreases and the noise increases under fog, rain, snow, and other weather conditions \cite{li2020happens,li2022common, dong2023benchmarking}. Especially, in the case of water, snow, etc. on the road, the LiDAR detection system may detect a large number of droplets or snow particles sprayed or splashed from the back and side of vehicles, causing the LiDAR perception model to judge it as phantom obstacles, reducing the average precision.
Therefore, it is particularly important to make up for the defects in the point cloud data under adverse weather conditions through the autonomous driving perception algorithm and enhance the robustness of the perception algorithm.

As the demand for both the quantity and quality of datasets continues to rise, data augmentation methods of adverse weather are continuously emerging \cite{hahner2020quantifying, dreissig2023survey}, primarily involving techniques such as effect into existing datasets or generation from simulated environments.
Some previous research has explored simulation methods for camera data, collecting RGB data under adverse weather conditions through simulation or generation, with the goal of minimizing domain gaps as much as possible \cite{sakaridis2018semantic, tremblay2021rain, Mirza_2022_CVPR}. 
Recently, there have also been studies augmentation adverse weather conditions for LiDAR data, proposing methods to simulate laser radar sensors in rain \cite{heinzler2019weather,goodin2019predicting,kilic2021lidar}, fog \cite{hahner2021fog, liu2022parallel} and snow \cite{hahner2022lidar}.
However, these methods cannot generate weather-affected data arbitrarily, as most of them rely on post-processing algorithms based on physical principles using real point cloud data captured in clear weather conditions as input. Additionally, they are unable to replicate dynamic phenomena associated with specific weather conditions, such as splashing effects.

\begin{figure}[!t]
   \centering
   \includegraphics[width=1.0\linewidth]{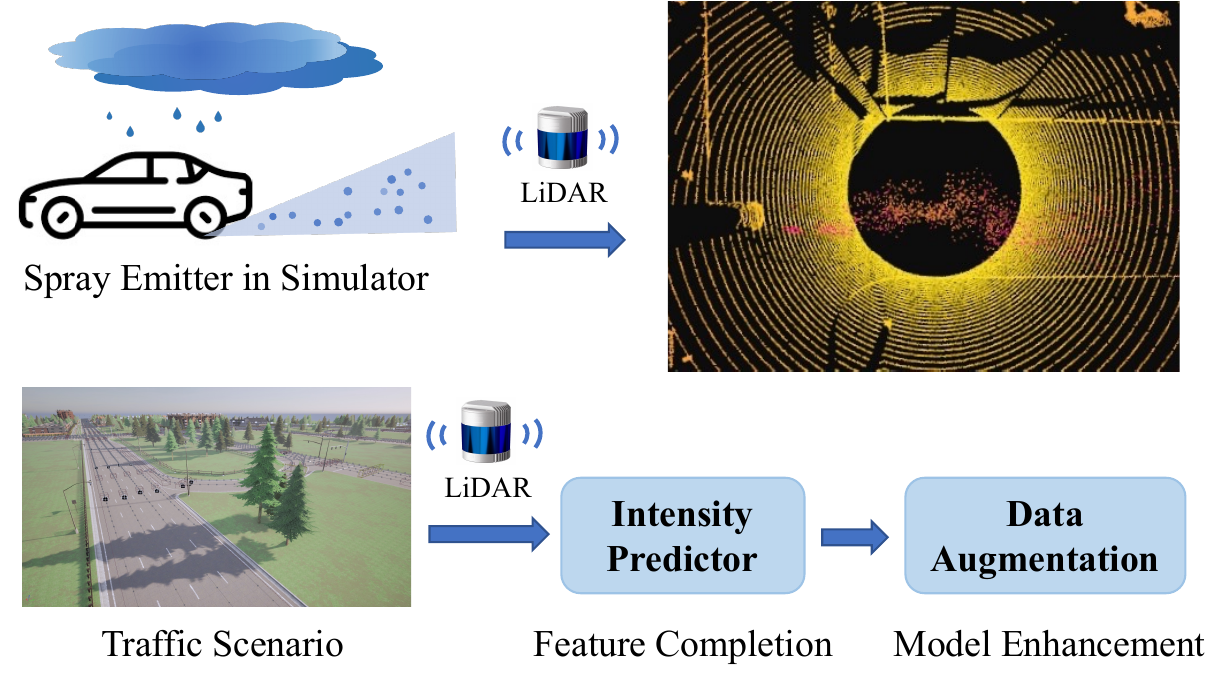}
   \caption{
       In our simulation environment, we model the spray phenomenon using a spray emitter and collected point cloud data with weather effects. We establish a data collection pipeline for rainy highway scenarios, where we enhance point cloud features using the Intensity Predictor. This process significantly contributes to data augmentation for 3D object detection tasks in adverse weather conditions.
   }
   \label{fig:header}
\end{figure}
\vspace{2mm}
To address these challenges, this paper presents a simulation method for LiDAR data under rain conditions, as shown in Figure \ref{fig:header}. We model the dynamic scene of rainy weather in the CARLA simulator \cite{dosovitskiy2017carla} and directly generate large-scale point clouds with rainy weather noise. Specifically, we simulate the presence of water on the road surface during rainy weather and the spray phenomenon occurring behind vehicles traveling at high speeds. However, due to the influence of object materials, the simulator cannot precisely calculate LiDAR echo intensity. To generate 4D point cloud data, we propose a weather-based point cloud intensity prediction network. This network calculates the intensity of LiDAR echo points based on object RGB material information, semantic label information, and weather information.

We develop a complete data generation pipeline to obtain high consistency between the synthetic dataset and the real dataset of the target domain while ensuring the generation efficiency, for data augmentation under adverse weather.
As we generate data in the simulator, we have the flexibility to use dynamic simulation models for a more accurate representation of various scenarios. Additionally, we can perform real-time rendering and collect sensor data with multiple parameter configurations.

Our contributions are summarized as follows:
\begin{itemize}
    \item We complete the modeling task of a rainy weather scene in the CARLA simulator, and establish the pipeline of detecting raindrops with LiDAR. Specifically, we pay special attention to the splashes and sprays caused by the wheels of surrounding vehicles during rainy weather, and simulate this phenomenon through our developed `Spray Emitter' method.
    \item We study the influence of different weather conditions on the intensity of the echo, establish a prediction network for the intensity of the LiDAR echo and complete the simulation of the 4-feat LiDAR point cloud.
    \item
    We introduce a synthetic point cloud dataset consisting of a total of 10,000 frames, encompassing data from various rainfall highway scenarios. We utilize the dataset for data augmentation in the 3D object detection task and observe that it effectively enhances the model's robustness to different weather conditions.

\end{itemize}

\section{Related Work}

\subsection{LiDAR Simulation in Adverse Weather}
LiDAR simulation in adverse weather aims to generate extra large-scale point cloud data that is challenging to collect and annotate in the real world to enhance models from a data-driven perspective 
To counteract this, work has been done to simulate LiDAR data under rainy, snowy, and foggy weather \cite{yang2020lanoising,hahner2021fog, hahner2022lidar,goodin2019predicting,kilic2021lidar}.  to make the perception models more robust in varying weather conditions, despite shortcomings in the original training dataset.
Yang \textit{et al.} \cite{yang2020lanoising} published a method to simulate foggy conditions within the collected LiDAR data. Moreover, Hahner \textit{et al.} \cite{hahner2021fog, hahner2022lidar} developed physically based models in the power signal domain for fog and snowy conditions and improved 3D object detection performance through data augmentation to expand the dataset. Goodin \textit{et al.} \cite{goodin2019predicting} present a simplified model to simulate the LiDAR performance in rain, while Kilic \textit{et al.} \cite{kilic2021lidar} present a comprehensive approach for adverse weather conditions based on Mie’s theory. 

In contrast to previous post-processing methods that augment existing datasets from clear conditions, our approach employs a simplified model and establishes a dedicated pipeline for the direct acquisition of synthetic point cloud data. Leveraging the simulator's advantages, we can reproduce dynamic weather phenomena and obtain precise labeling information.
However, one critical challenge we address is the domain gap between the simulation environment and the real world. To mitigate this gap, we employ a realistic LiDAR model and align the detail of the simulated scene with the real scene.
\begin{figure*}[t] 
\centering
\includegraphics[width=1.0\linewidth]{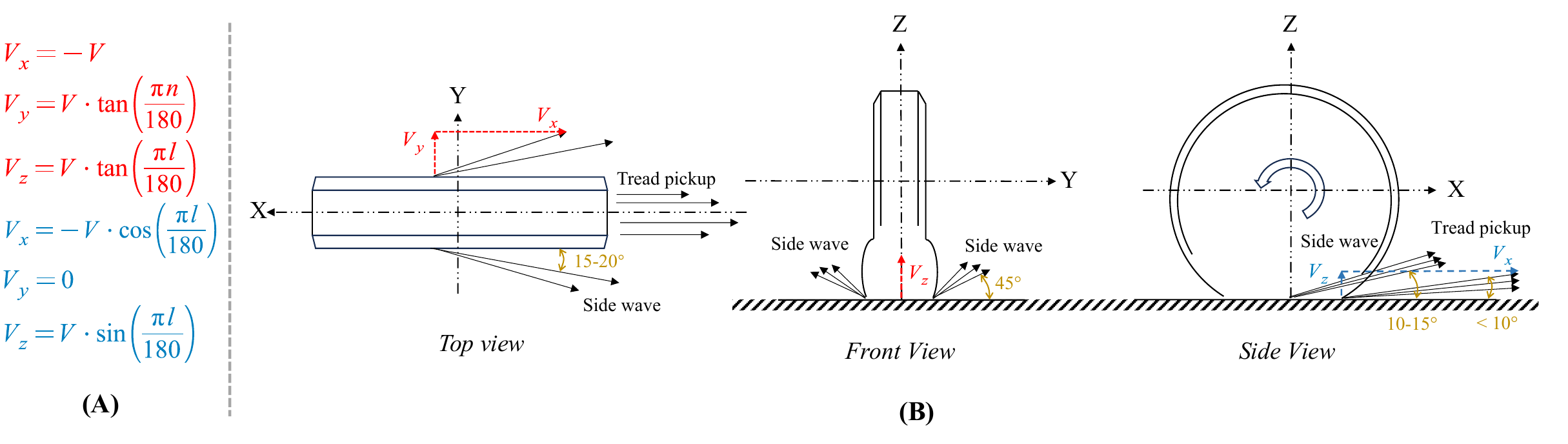}
\caption{\textbf{(A)} Calculation of velocity components projected onto Cartesian coordinates where $n$ represents the angle between the velocity projected onto the XOY plane and the negative X-axis, while $l$ represents the angle between the velocity projected onto the XOZ plane and the positive X-axis. \textbf{(B)} Spray Phenomenon Mechanism Analysis through top view, front view, and side view. The velocities of tread pickup is \textcolor{blue}{blue} and side wave is \textcolor{red}{red}.}
\label{fig:spray_mechanisms}
\end{figure*}
\subsection{Spray and Splash Effect Simulation}

Spray simulation is receiving increasing attention in the pursuit of autonomous driving due to the substantial noise it introduces to LiDAR data. Rivero \textit{et al.} \cite{rivero2021effect} presents a real-time simulation of the spatial distribution of detection on spray effect. 
Other approaches utilize physical models of spray to conduct simulation.
The early research was established by Weir\cite{weir1978reduction}, which categorized the generation of spray phenomenon into four mechanisms. Shih \textit{et al.} \cite{shih2022reconstruction} reconstructs spray patterns and synthesizes spray data by trajectory simulation and matching algorithms.  Linnhoff \textit{et al.} \cite{linnhoff2022simulating} proposed a clustering method for spray generation, and analyzed the disturbance pattern from real data. They also evaluate the improvement of LiDAR-based detectors trained with artificial spray data.
In comparison to the previous work, we utilize a physical model to simulate spray within the simulator, employing blueprints to dynamically generate spray particles in real time.
\subsection{Point Cloud Intensity}
The point cloud intensity captured by LiDAR constitutes a vital aspect of 3D detection and segmentation. LiDAR intensity is influenced by the geometric, physical, and environmental attributes of the LiDAR system. Dropout noise, resulting from light absorption and scattering, significantly impacts the quality of LiDAR data, with notable consequences for 3D perception. 
Squeezesegv2 \cite{wu2019squeezesegv2} proposed a point cloud intensity simulation method to obtain a dataset from GTA.
In a related development, Manivasagam \textit{et al.} \cite{manivasagam2020lidarsim} also employed U-net architecture to output a ray drop probability, and framed the LiDAR ray drop problem as a binary classification task. Vacek \textit{et al.} \cite{vacek2021learning} proposed a method to model intensity using a combination of LiDAR geometry, RGB images, and semantic labels. 
Compared with the previous works, Our primary focus is on the study of intensity prediction under varying weather conditions. 

\section{Physics Based Point Cloud Simulation in Rainy Weather}
\subsection{Spray Amount Calculation} 
In the stagnant water scene caused by heavy rain, the mechanisms of spray water caused by flying tires can be divided into 4, including bow wave, capillary adhesion, tread pickup, and side wave \cite{weir1978reduction}. Due to the obstruction of the vehicle chassis and tire muddy plates, the droplets caused by bow wave and capillary adhesion will quickly vanish due to collision. 
The tread pickup droplets are generated from the stagnant water attached to the tire gully. When the vehicle is running, the water in the tire drainage groove will rotate with the wheel at high speed, and finally spray out in the form of spray water on the rear side of the tire. 
Meanwhile, side wave may result in anomalies in vehicle point cloud contours. 
Therefore, we consider tread pickup and side wave as our primary areas of research, as shown in Figure \ref{fig:spray_mechanisms}. When LiDAR is operating, tread pickup droplets and side wave will form a barrier between the laser rays and the target, generating a large number of noise point clouds, reducing detection accuracy. 

The amount of jets of spray water is mainly related to the speed of the vehicle and the depth of the water on the ground. When the depth of the stagnant water is lower than the depth of the tire gully, the amount of water attached to the tire gully increases with the increase of the depth of the stagnant water until reaching the depth of the tire gully. When the speed is increased, the jet frequency of tread pickup will become higher and higher, resulting in a more significant phenomenon of spray water. The relationship between traffic and water volume and water volume can be summarized as described in the following formulas:
\begin{equation}
  \begin{aligned} \label{eqn-1}
    &VR_{TP} = Kbvh_{groove} \\
    &VR_{SD} = 0.5bv(WD -Kh_{groove} - (1-K)h_{film}) \\
    &WD = 6e^{-4}T^{0.09}(LI)^{0.6}S^{-0.33} 
  \end{aligned}
\end{equation}
where $VR_{TP}$ and $VR_{SD}$ respectively denote the volume rate of each mechanism.
$K$ is the groove width proportion of the tire.
$h_{groove}$ is the depth of water in the tire tread, $h_{film}$ is the depth of water picked up on each rotation of the tire.
$WD$ is the water film thickness, calculated from an empirical formula\cite{weir1978reduction}, where $T$ is the road texture depth, $L$ is the drainage length, $I$ represents rain rate and S represents slope.

\begin{figure}[t]
\centering
\includegraphics[width=1\linewidth]{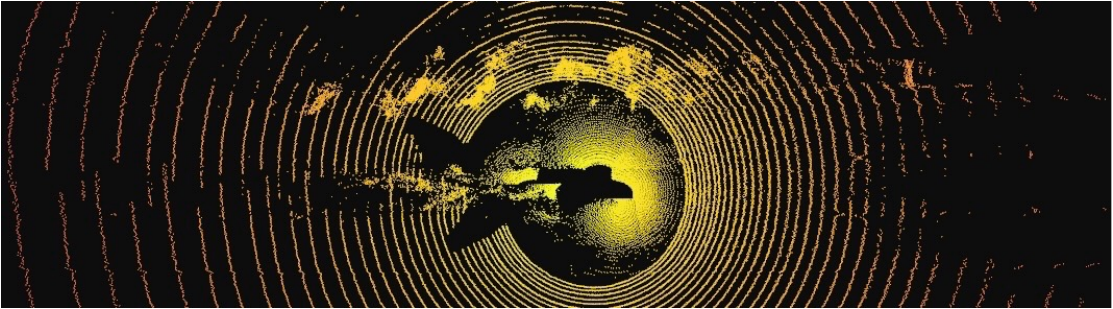}
\caption{Visualization with spray phenomenon from Waymo dataset.}
\label{fig:spray_demo}
\end{figure}

However, the depth of the water on the ground is not constant, and the uneven ground will lead to an increase in the randomness of the water depth. From Waymo Open Dataset, we can see that in Figure \ref{fig:spray_demo} due to the potholes on the ground, the tire splash phenomenon has a truncated effect. As a result, a well-defined cloud of water mist is formed at the rear of the speeding vehicle, rather than a continuous spray. We add a time-varying weight to the water volume in the range of 0.5 to 1.5 to simulate this effect. 

The spray droplets are different in size. General statistical experiments \cite{hasirlioglu2017introduction, gaylard2017surface,gaylard2011simulation,hagemeier2011practice} show that most of the droplet sizes in spray water range from 0.2 mm to 6 mm. The average value is about 1 mm. In this size, the shape of the liquid droplets is mostly a perfect sphere. In order to facilitate simulation, we use the classic value of 1 mm of a droplet size as the size of all spray droplets in the proposed simulation method and assume that the droplet is a perfect sphere.

As mentioned above, we reasonably approximates the phenomenon of spray water as tread pickup and side wave. When simulating the launch process of droplet particles, we separately calculated the particle emission quantities for the two mechanisms using the volume flow rate formula.

\subsection{Spray Droplets Dynamic Model}

\begin{figure}[tb]
\centering
\includegraphics[width=1.0\linewidth]{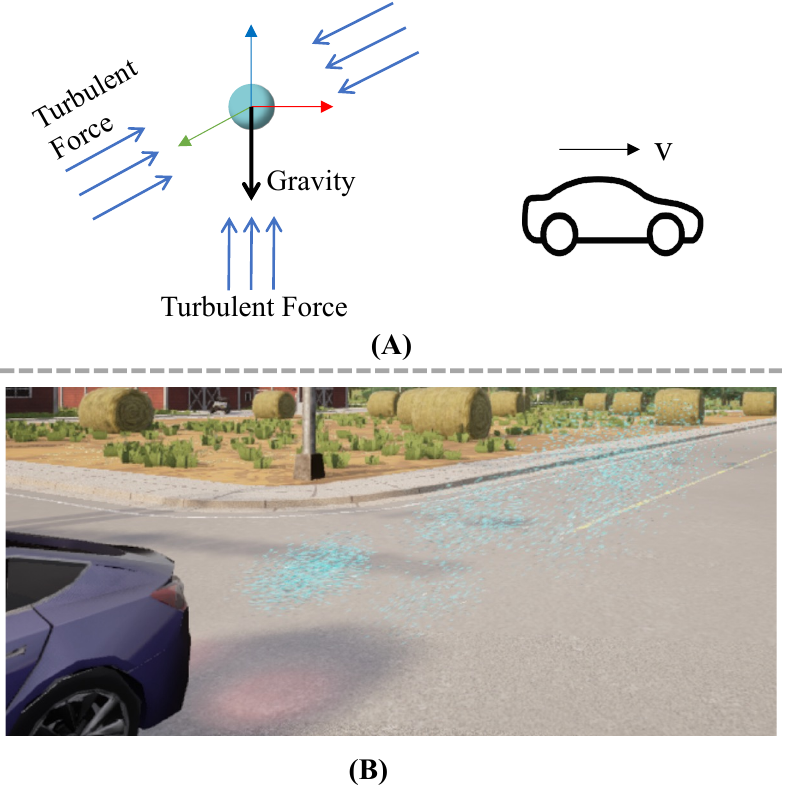}
\caption{(A) Force diagram of droplet. We neglect air resistance and do not depict the effect of crosswind on the droplets in the diagram. (B) Side view of spray effects in CARLA.}
\label{fig:ue_force}
\end{figure}
\vspace{1mm}
When the droplets are out of the tires, they will be influenced by the wind, gravity and the turbulence at the rear of vehicle, as shown in Figure \ref{fig:ue_force} (A).
Affected by the wake flow of a vehicle with high speed, spray or splash droplets swing along with the turbulence behind the vehicle. Yan \textit{et al.} \cite{yan2019experimental} analyzes the wake topology of the vehicle, which presents a bi-stable behavior. Airflow behind the vehicle is asymmetrical, while it is still statistically symmetrical over a long timescale. In response to this phenomenon, we adjust the mass flow rate of the two rear wheels, so that the spray and splash phenomenon can present an instantaneous asymmetric spatial distribution. Meanwhile, in order to simulate the spray droplets’ asymmetric trajectory, we introduced an initial lateral velocity for the droplets, whose direction is perpendicular to the forward direction of the vehicle. Then we apply a horizontal constraint to attenuate the lateral velocity. 
In addition to the turbulent force caused by the wake flow of the vehicle, during the flight of the droplet, it will also be affected by the crosswind of the road, gravity, air resistance, and so on.
The superposition of multiple forces will continuously change the motion state of the spray water in the air, forming an indeterminate particle group, to simulate its aerodynamic effect more realistically. 
According to the number of droplets emitted per frame, the emission speed calculated and the force analysis above, we generate spray water that is emitted as shown in Figure \ref{fig:ue_force} (B).
\subsection{Triggers for Spray Droplet Annihilation}

After detaching from the tire, the spray droplets annihilate anytime they collide with any object. In order to truly simulate the annihilation process of the spray and reduce the computing consumption as much as possible under the premise of ensuring the authenticity of the simulation, we design the following annihilation logic for the spray droplet:

\begin{enumerate}

\item When the spray droplet collides with any actor in the simulation environment, annihilation is triggered.
\item When the distance between the spray droplet and the LiDAR exceeds 75m, the effective detection range of Waymo LiDAR, the spray droplet has a weak impact on LiDAR detection, annihilation is triggered.
\item When the suspension time of the spray droplet exceeds 1.5s, we consider that it has broken or spread in the air and cannot be observed by LiDAR. Annihilation will be triggered at this time.
\end{enumerate}

\section{LiDAR Intensity Prediction}
\begin{figure}[tb]
\centering
\includegraphics[width=1.0\linewidth]{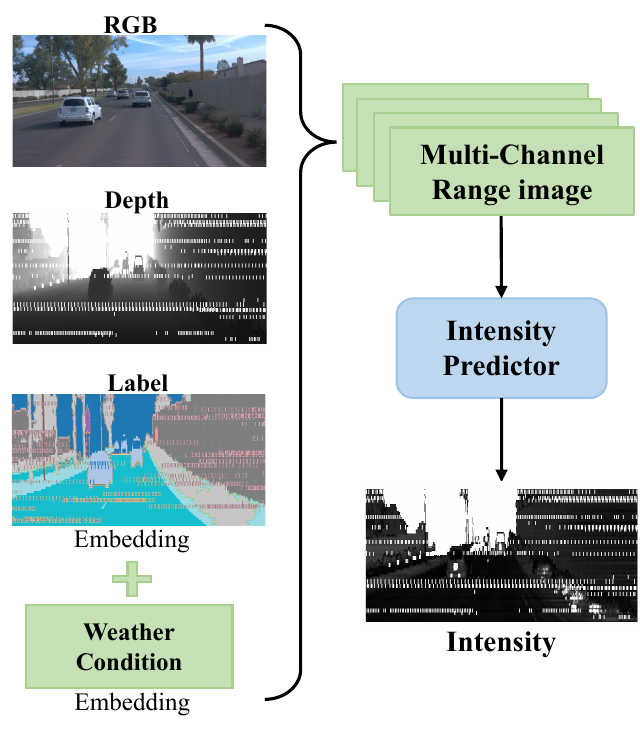}
\caption{Intensity Predictor: Render
the intensity from multi-channel input.}
\label{fig:predictor_pipeline}
\end{figure}
\subsection{LiDAR Intensity Principle}
In rainy conditions, the echo intensity of a LiDAR system undergoes significant variations due to the presence of raindrops and fog, which substantially alter the propagation and reflection characteristics of the laser beam. These factors are particularly critical when simulating LiDAR point clouds under adverse weather conditions like rain. We model the echo intensity to account for these variations.

In the point cloud data collected by LiDAR, the intensity is mapping the strength of the return power with the LiDAR receiver. The LiDAR back wave equation generally represents:

\begin{equation}\label{eqn-2}
  \begin{aligned}
    P_r & =  P_t  \Omega \rho \eta_{sys} \eta_{atm} \\
    \Omega & = \frac{\pi D_{rec}^2}{4R^2}
  \end{aligned}
\end{equation}

where $P_r$ is the returned laser energy.
$ P_t $ is the transmitted power, $\rho$ is the reflectance of the target surface.
$\Omega$ is the scattering steradian solid angle of the target. $\rho$ is the reflectance of the target surface, $\eta_{sys}$ the efficiency values of the optics of the system and $\eta_{atm}$ is the atmospheric attenuation. $D_{rec}$ represents the diameter of LiDAR receiver,  and $R$ is the distance of target.

However, directly obtaining object reflectance and atmospheric attenuation from the simulation environment is not feasible. Therefore, we present a weather condition-based intensity predictor.

\subsection{Intensity Prediction Neural Network}
Following the approach proposed by \cite{wu2019squeezesegv2, vacek2021learning}, we utilize a U-net \cite{ronneberger2015u} neural network to learn point cloud intensity from real data. Based on the descriptions in the formula, we estimate object reflectance from camera RGB data and LiDAR semantic label information. We combine this with LiDAR depth information to generate point cloud intensity. However, estimating LiDAR point cloud intensity under the influence of weather remains challenging. Hence, we include additional weather conditions as an input to the intensity predictor, as illustrated in Figure~\ref{fig:predictor_pipeline}. Specifically, we consider the combined influence of multiple factors, using RGB images, semantic labels, depth, and weather conditions as inputs. 

All the data within the intensity predictor are represented in the LiDAR Spherical coordinate system and organized in the form of a multi-channel range image. 
For the semantic information of the LiDAR point cloud, we directly obtain it from the Waymo dataset. To facilitate subsequent inference on the synthetic dataset, we align the semantic categories with the semantic labels in the CARLA simulator.
Regarding the acquisition of rainy information, we annotate rainy conditions at the segment level in the Waymo dataset, categorizing all segments into four types: clear, wet ground, light rain, and heavy rain. We convert the discrete inputs into the predictor using Embedding layers.
Same as \cite{vacek2021learning}, we project the sparse LiDAR point cloud onto the five camera images in the Waymo dataset to obtain RGB information. In addition, we have introduced the RGB mask and ray drop-off mask in order to compute the loss. We supervise the intensity predictor by masked L2 loss.
\begin{equation}\label{eqn:loss}
  \begin{aligned}
    \mathcal{L}_2=\mathbf{MSE}(\hat{I}, \ I) \cdot M
  \end{aligned}
\end{equation}
where $\hat{I}$, $I$ donate the predicted intensity range image and ground true intensity range image and $M$ is the union of RGB mask and drop-off mask.


\subsection{Model Validation}

\begin{table}[ht]
\small
\centering
\resizebox{0.99\linewidth}{!}
{
\setlength{\tabcolsep}{1.6mm}{
\begin{tabular}{c|c|c|c|c}
\hline
\multirow{2}{*}{Input Channel} & \multirow{2}{*}{Range Size} & 
\multicolumn{3}{c}{Intensity RMSE} \\
\hhline{~~|-|-|-}
& & $\text{All}$ & $\text{Light Rain}$ & $\text{Heavy Rain}$ \\
\hline
RGB+D+L & $64 \times 640$ & 0.0871 & 0.0721 & 0.0685  \\
RGB+D+L+W & $64 \times 640$ & 0.0848 & 0.0721 & 0.0700 \\ 
\hline
RGB+D+L & $64 \times 1376$ & 0.0948 & 0.0830 & 0.0781 \\ 
RGB+D+L+W & $64 \times 1376$ & 0.0911 & 0.0793 & 0.0774 \\ 
\hline
\end{tabular}}
}
\caption{Additional weather input channel results using intensity RMSE, where All donates the result on all weather conditions including sunny, light rain, and heavy rain.}
\label{tab:predictor_result}
\end{table}
We conducted an assessment of the Intensity Predictor using the root mean squared error (RMSE) metric, as shown in Table~\ref{tab:predictor_result}. All experiments shared the same setup, including an initial learning rate of 0.01, the utilization of a cosine decay learning strategy, and a weight decay value set at 0.01. We evaluate the impact of additional input of weather condition channel and different width of range images on the Intensity Predictor. It's evident that the addition of weather condition data significantly reduces the predictor's errors, with errors decreasing as precipitation intensity increases. Similarly, we conduct tests with a larger viewing angle range (approximately 180 degrees) and observe that the input of rainy conditions has a more pronounced effect on the model.

\section{Experiments}
\subsection{Point Cloud synthetic dataset in Adverse Weather}

We leverage the realistic self-driving simulator CARLA \cite{dosovitskiy2017carla} to implement the above algorithm. 
Specifically, we utilize the blueprint visual scripting system in UE4 to construct the simulated scene. We add a “splash emitter” actor component to the base vehicle blueprint in the simulator. By dynamically acquiring various information from the simulated environment, such as the position of the vehicle's rear wheels, vehicle speed, map data, and weather parameters, the “splash emitter” uses this information to generate and control the splash droplets. This ultimately achieves high-quality, dynamic simulation of the phenomenon of splash droplets in the real world.
In order to reduce computing resource usage, we revise the modeling of droplets. We divide a single group of droplets into a cluster consisting of the central droplet and nine surrounding droplets, randomly distributed within 5mm of the central droplet. The entire group of ten droplets functions as a single actor in the simulation framework. This allows us to decrease the number of actors by a factor of ten, albeit with a slight sacrifice in simulation accuracy.
In the autonomous driving simulation platform, in order to achieve smaller domain gaps between the synthetic data and real-world Waymo data, we use the RLS Library framework \cite{10161027}. We extend Waymo LiDAR to the RLS Library framework and complete the simulation of the Waymo LiDAR physical model in the simulation engine. According to the LiDAR information described by Waymo Open Dataset \cite{sun2020scalability}, it shows that Waymo LiDAR has the following physical characteristics,
\begin{enumerate}
\item The type of Waymo LiDAR is mechanical rotating LiDAR.
\item The number of lasers emitted per second is about 1.6 million.
\item The maximum effective detection distance is 75 m.
\item The vertical field of view ranges from -17.6 to +2.4 degrees.
\item The placement position is 2 meters over the ground and on the central axis of the vehicle.
\end{enumerate}

Additionally, we statistically analyze the point cloud intensity distribution in the Waymo Open Dataset \cite{sun2020scalability}, and finally calculate the point cloud drop ratio of the Waymo LiDAR.

In the selection of simulation scenarios, we opt for highway scenarios that are more susceptible to the effects of rainfall. 
In conventional urban roads, the impact of heavy rain on LiDAR is mainly from the droplets in the landing.
In the highway section, due to the high speed of vehicles, the phenomenon of Spray or Splash Water becomes more common and fierce, which is more challenging to LiDAR perception. 
During the generation of the synthetic dataset, we select the highway section of the map Town07 as the primary simulation scenario.
By dynamically controlling traffic flow, we ensure that there are constantly 1-6 other vehicles driving around the ego vehicle, ensuring that the sensors on the ego vehicle can collect sensor data with spray effects. In this scene, the vehicle speed ranges from 80km/h to 100km/h, and the amount of rainfall is from 30mm/h to 60mm/h. In most cases, the water depth in the area exceeds the 3.5mm depth of the tire groove.

We utilize the OpenCDA data collection framework \cite{xu2021opencda} to simultaneously collect LiDAR data, semantic LiDAR data, panoramic view of RGB images, and weather and vehicle labeling information in the traffic scene. We collect a total of 10,000 frames of data under various rainfall conditions at the frequency of 10 Hz.

After collecting synthetic data with rainy weather effects, we employ the intensity predictor to add the 4th feature to the point cloud data. 
Specifically, we partition synthetic data into front and rear 180-degree views. We then use the predictor trained on Waymo dataset's forward 180-degree data to make inferences on these segments, ultimately generating complete 4-feature point cloud data under varying rainy weather conditions.

However, due to the lack of annotated spray labels in the intensity predictor training data, the estimation of intensity for synthetic data with spray noise is inaccurate. Therefore, we conduct statistical analysis of point cloud data of the Spray droplets detected by LiDAR in the Waymo Open Dataset. We find that the intensity of the point cloud there is roughly in the range of 0.002 to 0.003. After the ray emitted from LiDAR hits the spray water, most of the energy is transmitted through the droplets. Only a small amount of energy can reflect, which becomes an echo and is eventually received by the LiDAR receiving. Therefore, when assembling point cloud intensity, we consider the point cloud intensity of spray noise points as random variables following a Gaussian distribution with a mean of 0.0025 and a variance of 0.0004.

\subsection{Experiment Setup}

After the data collection pipeline, we simulate 4-feat point cloud data under rainy weather conditions. In the dataset configuration, we directly merge the Waymo dataset with the synthetic dataset in adverse weather, aligning label domains and point cloud coordinate systems. We sample 1,600 frames of rainy synthetic data and 16,000 frames from the Waymo dataset, which are annotated as sunny weather, as the training set. Based on weather annotations, we sample 1,200 frames each for the real sunny weather test set and the real rainy weather test set. We also ensure that the training set did not contain the same sequences from the test set.

All experiments are trained from scratch and implemented with MMDetection3D\cite{mmdet3d2020}. For training on the directly merged dataset, we use Adam optimizer with a multi-step scheduler, where we set the initial learning rate to 0.001 and the weight decay to 0.01. The entire training process concludes upon reaching 24 epochs using 4 NVIDIA 3090 GPUs. We use AP for car category evaluation in both the Bird’s Eye
View (BEV) and 3D over 40 recall positions, where the IoU threshold is set to 0.5 and 0.7. 

\subsection{Experimental Results}

\begin{table}[ht]
\small
\centering
\resizebox{0.99\linewidth}{!}
{
\setlength{\tabcolsep}{1.6mm}{
    \begin{tabular}{c|c|c|c|c}
    \hline
        \multirow{2}{*}{Setting} &
        \multicolumn{2}{c|}{Test on Sunny} &
        \multicolumn{2}{c}{Test on Rainy} \\
        \hhline{~|-|-|-|-}
         & $\text{AP@0.5}$ & $\text{AP@0.7}$ & $\text{AP@0.5}$ & $\text{AP@0.7}$ \\ \hline
        Sunny\&Rainy & 79.21/76.45& 70.74/50.17& 79.14/76.05& 75.31/38.59 \\ \hline
        Sunny\&Sim& 79.18/\textbf{78.70}& 
\textbf{72.96}/\textbf{56.71}& \textbf{81.52}/\textbf{78.46}& \textbf{75.44}/36.39 \\ \hline
\end{tabular}}
}
\caption{Merged dataset results using Pointpillars \cite{Lang_2019_CVPR} baseline. Sunny represents a total of 16k frames Waymo sunny training set, Sim and Rainy represent a total of 1.6k frames synthetic rainy training set and Waymo rainy training set. The AP is represents as $AP_{BEV}$/$AP_{3D}$.}
\label{tab:merged_set_result}
\end{table}

To validate the authenticity of 4-feat synthetic point cloud data generated by our method and the effectiveness of data augmentation in downstream tasks, we conduct experiments by evaluating the model's performance trained on various merged datasets.
According to Table~\ref{tab:merged_set_result}, our model train on Sunny\&Sim merged dataset has a clear advantage. The model outperforms on rainy test set by around 2\% at AP@0.5 and also maintains accuracy on the sunny test set compared to the baseline. Experimental results confirm that our synthetic point cloud data closely approximates the distribution and characteristics of the Waymo real dataset, thereby validating the authenticity of our synthetic data. 
Furthermore, we believe that by increasing the proportion of synthetic data to a certain extent, the model's performance in rainy weather scenarios is expected to be better.
\vspace{1.5mm}
\section{Conclusions}
In this work, our primary goal is to achieve realistic simulation for LiDARs under rainy weather conditions.
We introduce a simulation approach based on physical models to simulate the dynamic effects of adverse weather conditions in a simulated environment, specifically, focusing on simulating the spray or splash effect. Furthermore, we reconstruct the traffic scenarios under rainy weather in CARLA and develop a weather-based intensity prediction network to generate point cloud intensities. This enables us to set up a data generation pipeline for creating realistic synthetic dataset. 
The experimental results from the perception model demonstrate that the synthetic data we generate exhibits a reduced domain gap, enhancing the model's robustness in rainy weather conditions.

\vspace{4.5mm}
\textbf{Acknowledgement.} 
The research was supported by the National Key R\&D Program of China (Grant No. 2022ZD0160104) and the Science and Technology Commission of Shanghai Municipality (Grant No. 22DZ1100102).









\newpage

\bibliographystyle{IEEEtran}
\bibliography{IEEEexample}

\end{document}